\pdfoutput=1

\documentclass[11pt]{article}

\usepackage{EMNLP2022}

\usepackage{times}
\usepackage{latexsym}

\usepackage[T1]{fontenc}

\usepackage[utf8]{inputenc}

\usepackage{microtype}

\usepackage{inconsolata}
\usepackage{nert}

\setlist{  
  listparindent=\parindent
}

\usepackage{adjustbox}
\usepackage{color, colortbl}

\newcommand{\deprel}[1]{\texttt{#1}}
\newcommand{\func}[1]{\textsf{#1}}

\definecolor{Gray}{gray}{0.9}

\title{CGELBank: CGEL as a Framework for English Syntax Annotation
}
\author{Brett Reynolds\\
Humber College\\
\eml{brett.reynolds@humber.ca} \And 
Aryaman Arora \quad Nathan Schneider\\
Georgetown University\\
\{\emldisplay{aa2190@georgetown.edu}{aa2190}, \emldisplay{nathan.schneider@georgetown.edu}{nathan.schneider}\}\texttt{@georgetown.edu}}
\date{}

\graphicspath{{trees/}}

\begin{document}

\maketitle

\begin{abstract}
We introduce the syntactic formalism of the \textit{Cambridge Grammar of the English Language} (CGEL) to the world of treebanking through the CGELBank project. We discuss some issues in linguistic analysis that arose in adapting the formalism to corpus annotation, followed by quantitative and qualitative comparisons with parallel UD and PTB treebanks. We argue that CGEL provides a good tradeoff between comprehensiveness of analysis and usability for annotation, which motivates expanding the treebank with automatic conversion in the future.
\end{abstract}

\section{Introduction}

Parsing hierarchical syntactic structure 
is a central endeavour in computational linguistics \citep{slp3,kubler2009dependency}. 
Many syntactic theories and annotation frameworks exist. The venerable Penn Treebank \citep[PTB;][]{marcus-etal-1994-penn} has been the leading approach to annotating English constituent structure; its detailed annotation guidelines have been applied to many corpora over the years \citep[e.g.,][]{ewt,ontonotes}.
Other theories applied to large-scale annotation for English have included Universal Dependencies \citep[UD;][]{nivre-etal-2016-universal}, Combinatory Categorial Grammar \citep{ccg}, Role and Reference Grammar \citep{bladier2018rrgbank}, Head-Driven Phrase Structure Grammar \citep{miyao2004corpus}, and so on. Each formalism makes different theoretical claims (e.g., are there transformations?)\ and computational tradeoffs (e.g.~complexity vs.~parsing efficiency).

\begin{figure}
    \centering
    \includegraphics[width=0.4\textwidth]{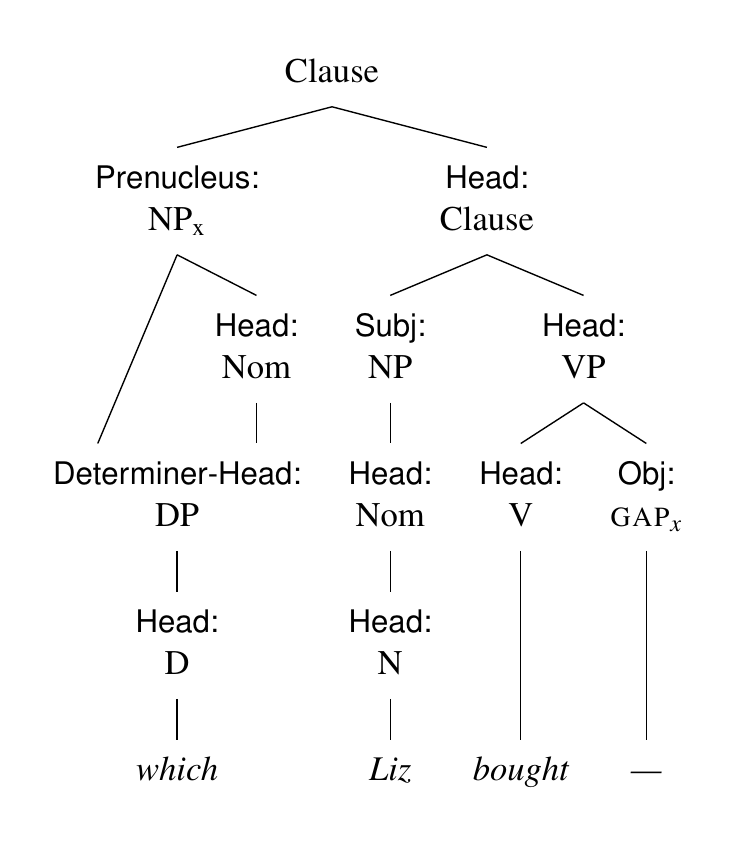}
    \caption{CGELBank-style tree for interrogative clause \pex{which Liz bought} (as in \pex{There are three options; I wonder which Liz bought}).}
    \label{fig:5brevised}
\end{figure}

In this paper, we introduce a treebank for English built on the syntactic formalism of the \textit{Cambridge Grammar of the English Language} \citep[\textbf{CGEL};][]{cgel}. CGEL is the authoritative descriptive grammar for English and analyses many syntactic phenomena in extreme detail with minimal theoretical claims (\cref{sec:why}). Our motivations for introducing yet another treebank are: (a)~existing annotation guidelines for other formalisms cannot approach the depth of CGEL; (b)~neither of the most common frameworks for descriptive annotation in English (PTB and UD) offers a coherent account of both constituent structure and grammatical functions; and (c)~annotating a real-world treebank is a strong test of the expressivity and consistency of the CGEL formalism.

Below, we delve into the linguistic issues surmounted in building the initial version of \textbf{CGELBank} (\cref{sec:decisions}).\footnote{\url{https://github.com/nert-nlp/cgel/}} 
We then compare the CGEL trees with UD and PTB 
trees of the same sentences (\cref{sec:comparison}).



\section{What is CGEL?}\label{sec:why}


CGEL is the most comprehensive and up-to-date descriptive grammar of English \citep{brew-03}. It uses a morphosyntactic formalism for describing English from first principles \citep{pullum2008expressive}, wherein the hierarchical structuring of spans as constituents is supplemented with labelling of the dependency-like grammatical functions between them.

CGEL posits a distributionally-defined set of nine lexical categories, from which we developed a part-of-speech tagset with 11 tags: N (noun), N\textsubscript{pro} (pronoun), V (verb), V\textsubscript{aux} (auxiliary verb), P (preposition), D (determinative), Adj (adjective), Adv (adverb), Sdr (subordinator), Coordinator, and Int (interjection). Pronouns and proper nouns are a subset of nouns,
though we have created a distinct tag for pronouns; auxiliary verbs are a subset of verbs. All of these categories except subordinator and coordinator 
project higher-level phrasal constituents, e.g.~N $\gets$ Nom (nominal) $\gets$ NP (noun phrase).

Each CGEL phrase has exactly one \func{Head} function along with zero or more dependents. There are also a few non-phrasal dependent constituents in flat structures (e.g.~\func{Coordinate}). Phrases are typically binary- or unary-branching, but $n$-ary branches are also possible. CGEL contrasts adjuncts (\func{Mod}, \func{Supplement}) with complements (\func{Comp} and subtypes shown in \cref{fig:FunctionSet} of \cref{sec:functions}). It also analyses many complex syntactic phenomena often ignored in computational work, e.g.~gaps (\cref{sec:gap}) and scope of coordination (\cref{sec:coord-gaps}).

Multilingual syntactic formalisms such as Universal Dependencies \citep[UD;][]{nivre-etal-2016-universal} lack the expressivity of CGEL for specifically English due to the necessity of cross-lingual consistency.\footnote{E.g., the general principle of content-heads, leading to verbal auxiliaries being treated as dependents of the main verb in English but main-verb copulae (which are not used in some languages) being analysed as dependents of the predicate.} CGEL also differs from the Penn Treebank  \citep[PTB;][]{marcus-etal-1994-penn} in associating grammatical functions to each edge between phrasal categories in a tree, adopting the concept of headedness from dependency grammars. CGEL also analyzes ``transformational'' phenomena in English with gaps, but in a less overwhelming manner than PTB, limiting their uses to ellipsis and relations that would otherwise be non-projective.

As a whole, we find that CGEL combines the best of both worlds---comprehensive analysis of complex and long-tail linguistic phenomena with the minimal parse complexity needed to express those, as well as both dependency and constituency layers.
And it does so in human-readable fashion, precisely defining its terminology and defending its analyses in a nearly 2000-page volume that is widely referenced by linguists. 
Finally, a companion textbook \citep{sieg1,sieg2} 
introduces language learners to the major points of English syntax.
A CGEL-style treebank (CGELBank), potentially with a parser, would therefore be of interest to English learners familiar with the framework.


\section{Linguistic decisions}\label{sec:decisions}
Despite its breadth, depth, and specificity \cite{Culicover2004}, there are elements of English grammar that are not fully specified in CGEL. While CGEL does use some corpus data in its analyses and descriptions of English, the grammar is largely based on contrived examples. CGEL also only describes an idealized standard variety of English---internet-sourced text is out-of-distribution. As a treebank forces us to be explicit in our linguistic decisions, here we describe a number of the issues we ran into and the (not necessarily final) decisions we made.

    \subsection{Categorizing individual lexemes}
    Designing part-of-speech (POS) tagsets and delineating boundaries between tags has long been a contentious problem in treebanking \citep{atwell2008development}.
    CGEL is detailed but non-exhaustive in this regard.
    
    In developing CGELBank, we had to collate the many mentions of lexemes and their categories distributed throughout CGEL,\footnote{E.g., ``One borderline case is \pex{else}. This is an adverb when following \pex{or}, as in \pex{Hurry up or else you’ll miss the bus}, but arguably a preposition when it postmodifies interrogative heads and compound determinatives.'' (CGEL, fn.~5, p.~15)} along with the careful application of CGEL principles to the categorization of hundredes of lexemes not explicitly mentioned.\footnote{ Carried out since 2006 in consultation with Huddleston and Pullum and recorded in the \href{https://simple.wiktionary.org/}{Simple English Wiktionary}. \finalversion{we could also cite Brett's paper on N\textsubscript{pro} and D in the camera ready}} Examples of words categorized this way include the determinative \pex{said} (e.g., \pex{as in \uline{said} contract}), the coordinator \pex{slash} (e.g., \pex{Dear God \uline{slash} Allah \uline{slash} Buddha \uline{slash} Zeus}), and the preposition \pex{o'clock} \citep{pullum-13}.
    
    \subsection{Simplifying and un-simplifying}
    As shown in CGEL's figure 5b (p.~48; reproduced in the appendix as \cref{fig:5b}), CGEL uses a variety of subtypes of head within clause structure: \func{Nucleus} is the head of a clause which is itself a clause, \func{Predicate} is a VP that is the head of a clause, and \func{Predicator} is the V that is the head of a VP. We dispense with these subtypes, simply using \func{Head} in all cases, as shown in \cref{fig:5brevised}. 
    
    

    In some cases, CGEL simplifies tree structures by removing intermediate unary nodes, such as by removing an intermediate \func{Head}:Nom between \func{Head}:N and its projected NP. We always include such nodes, as in \cref{fig:5brevised}.
    Moreover, we show complements as sisters of N and V but modifiers as sisters of Nom and VP, where CGEL again simplifies at times.
    

    \subsection{Gaps}\label{sec:gap}
    CGEL posits gaps in tree structures when a constituent appears in prenucleus position, as in \pex{which Liz bought} in \cref{fig:5brevised}, but is inconsistent in indicating it. In most cases, we have decided to explicitly indicate a gap. We identify the following unclear cases and present our decisions.

        \subsubsection{Subject gaps}
        In open interrogatives such as (\ref{ex:interrogative}a), ``inversion accompanies the placement in prenuclear position of a non-subject interrogative phrase'' (CGEL, p.~95), while there is no inversion in those like (\ref{ex:interrogative}b). 
        
        \ex.\label{ex:interrogative} 
            \a. What did she tell you?
            \b. Who told you that?
        
        This suggests two possible analyses for the structure of (\ref{ex:interrogative}b): either they both have a prenucleus and a co-indexed gap, or only (a) does, the \pex{who} in (b) being a normal subject
        \cite{Maling2000}. Unfortunately, it is not clear to us which position CGEL takes. The discussion on p.~96 suggests that there is no inversion in (\ref{ex:interrogative}b) \textbf{because} there is no prenucleus, and thus no gap. However, this is not a consistent rule throughout the text.%
    %
        %
        \footnote{CGEL recognizes explicitly that a subject gap may exist in a construction like \pex{\uline{Who}}\textsubscript{i}~[\pex{do~you think}~[\pex{\uline{~~~}}\textsubscript{i} was responsible]]\pex{?}\ (p.~1082), but this involves a subordinate clause.}
       Given the ambiguity, we have taken what we see as the standard position that there is indeed a gap \citep[e.g.][]{Maling2000,bies1995bracketing} in clauses with subjects that have been questioned or relativized.
        
        
        \subsubsection{Adjunct gaps}
        The issue here concerns adjuncts such as the PP in \pex{after lunch, we left} and whether or not they are co-indexed to a VP-final gap. CGEL recognizes that ``adjuncts may also be located in prenuclear position'' (p.~1372), but ``there need be no anaphoric link with a pronoun [or gap] in the nucleus of the clause'' (p.~1410), as in \pex{\uline{As for the concert-hall}, the architect excelled herself}, where the underlined section is in prenucleus position but cannot appear clause finally. Other examples of items appearing in pre- or postnuclear position without a corresponding gap include the relative clause in an \pex{it}-cleft like \pex{It's the director \uline{who was sacked}} and markers (e.g., \pex{\uline{whether} it works}, p.~956).
        
        Because adjuncts appear in a variety of locations in clause structure, with some not appearing clause finally, we have decided against including a gap.
        The exception is in relative and open interrogative clauses, where CGEL explicitly marks a gap (e.g., 
        \pex{That’s not the reason} [\pex{why}\textsubscript{i}~[\pex{he~did it~\uline{~~~}}\textsubscript{i}]], p.~1086). 
        
        \subsubsection{Phrasal genitives}
        CGEL has the genitive suffix \pex{'s} attaching to the last word in an NP such as [\pex{somebody local's}]. In the case of an NP ending in a gap, we take the \pex{'s} as attached to that gap, as in \pex{a guy I know~\uline{~~~}'s~house}.

            
        \subsubsection{Coordination and comparatives} \label{sec:coord-gaps}
        CGEL calls coordinations such as \pex{\uline{The PM} arrived \uline{at six} and \uline{the Queen} \uline{~~~} \uline{an hour later}} \func{Gapped Coordination}, and writes examples like this with a ``gap'' as shown here between \pex{the Queen} and \pex{an hour later} (p.~1338). But this is not the same kind of gap we find in long-distance dependencies such as the subject relatives discussed above; rather, this is ellipsis.
        Consequently, CGELBank does not include a gap in tree structure. Instead, we have a nonce coordinate NP + PP with daughters \func{Subj}:NP + \func{PredComp}:PP or the like. Similarly, we do not treat the ``gaps'' in comparatives as gaps in tree structure.
        
        
    \subsection{Branching \& tree structure}

        \subsubsection{Extraposition}\label{sec:extraposition}
        CGEL posits extraposed subjects and objects, such as the underlined clause in \pex{It's a good thing \uline{that we left early}},\footnote{These are semantic agents, patients, etc., but not syntactic \func{Subj} and \func{Obj}.} saying that its extraposed subject is ``in a matrix clause containing \pex{be} + a short predicative complement'' (p.~953). Unfortunately, this leaves the precise structure unclear.
        
        
            \label{fig:2naryExtraposition}
        
            \label{fig:SEIG2extraposition}
        
        After considering various options, we have decided to attach the extraposed constituent as a second complement in the VP with ternary branching.
        Despite the seeming difference implied by the labels ``extraposed'' and ``internalized'', we see this as analogous to the position of the internalized complement---the \pex{by} phrase---in the VP in passive clauses  (p.~674).

        
      
        

        \subsubsection{Coordinates \& markers}\label{sec:coord}
\begin{figure}
            \centering
            \includegraphics[width=\columnwidth]{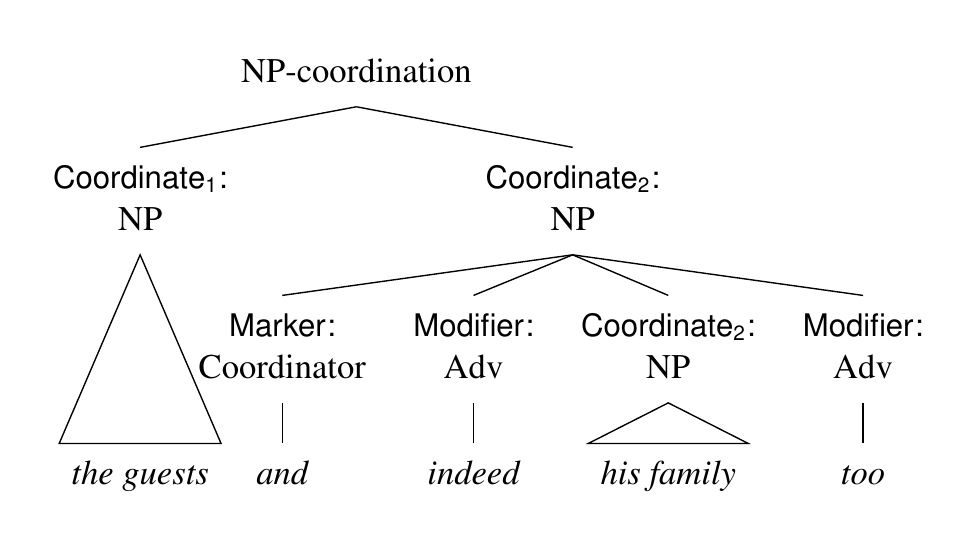}
            \caption{CGEL flat coordination---rejected in CGELBank, where \pex{indeed his family too} is an NP serving as the \func{Head} of the second coordinate.}
            \label{fig:Coordination}
        \end{figure}

        A coordi\textbf{nation} is a non-headed construction with coordi\textbf{nates} as daughters (CGEL  p.~1278).\finalversion{\footnote{Unlike CGEL, we include the functions (e.g., \func{Obj} + \func{Mod}) in nonce-coordination.\nss{I don't follow---an example would help. Maybe save for the camera-ready}\br{sure}}} Therefore, coordi\textbf{nates} are neither heads nor dependents. Consider, though, the following coordination \pex{the guests and indeed his family too} (p.~1278), 
        reproduced here as \cref{fig:Coordination}.
        
        Unlike coordinations, NPs like \pex{and indeed his family too} are headed constructions. The NP has two modifiers: \pex{indeed} and \pex{too}, which, like all modifiers, are dependents requiring a head sister. But if coordinates are not heads, then this NP is headless.
        
        \func{Marker}s\footnote{Though CGEL uses ``marker'' both non-technically (e.g., marker of distinctively informal style), and technically as a function term, we discuss only the latter.} are sisters of heads when they are subordinators (see (9) on p.~954 and (51) on p.~1187), so a marker is a dependent. This, however, is incompatible with \func{Marker}:Coordinators like \pex{and} in \pex{and indeed his family too} being the sister to the \func{Coordinate}:NP \pex{his family} (p.~1277).
        
        Given these facts, the NP \pex{his family} in \cref{fig:Coordination} must be a head and not a coordinate. We generalize from this to the principle that a coordinate is only the daughter of coordination, and a marker is always a dependent with a sister head.

        \subsubsection{Indirect complements} \label{sec:IndirectComp}
\begin{figure}
            \centering
            \includegraphics[width=\columnwidth]{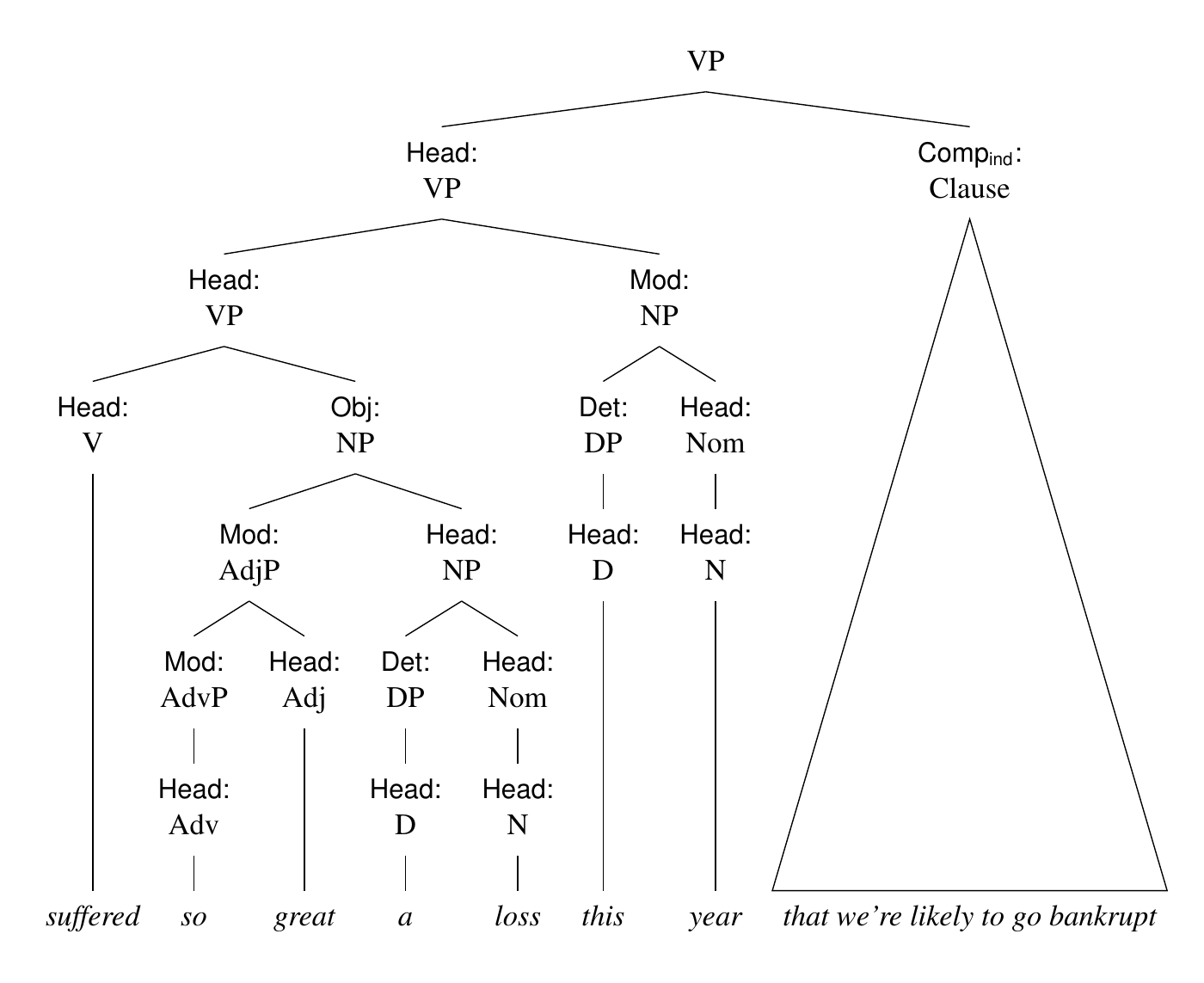}
            \caption{CGELBank structure of the indirect complement of \pex{so}.}
            \label{fig:CompIndLong}
        \end{figure}

        Indirect complements are those such as the underlined phrase in \pex{enough time \uline{to complete the work}} which are licensed by a dependent in the phrase, here the \func{Det} \pex{enough}, but CGEL is inexplicit about it attachment. We construct a superordinate phrase of the head type and branch the indirect complement from that, as in [[\pex{enough time}]\textsubscript{\func{Head}:NP} [\pex{to do the work}]\textsubscript{\func{Comp\textsubscript{rel}}:Clause}]\textsubscript{NP}. When the complement is further delayed, we branch it from the nearest possible parent phrase, as in \cref{fig:CompIndLong}.
        
        
        \subsubsection{Verbless clauses}
        CGEL's treatment of so-called ``verbless clauses'' (VlCs; ch.~14§10) is incompatible with its treatment of clauses in general. Though CGEL provides no definition of a clause, a clause is a phrase (p.~23) and a phrase is a constituent with a head and possible dependents (p.~22). A predicate is a VP that functions as the head of a clause (p.~24). Therefore, a clause is a phrase headed by a VP. But VlCs have no verb and no VP, so they must not be clauses in the syntactic sense that CGEL implies. That is not to deny that, like clauses, VlCs express a semantic connection between a predicand and a preciate, but there are other constructions in CGEL that do this without being analyzed as clauses (e.g., \pex{it made \uline{them happy}}; p.~217).
        
        
        Seeing no clear alternative, we treat PPs like \pex{with his hands in his pockets} as having two complements, analogous to complex transitive verbs (e.g., \pex{made him happy}) and PPs like \pex{while happy} as taking predicative complements, similar to \pex{as}. Finally, we analyze supplement VlCs (e.g., \pex{\uline{bag in hand}, they set out}) as headless nonce constructions like those in gapped coordination (see \cref{sec:coord}).
        \finalversion{\nss{Doing away with the concept of VlC's may surprise some people. Aren't they sort of like zero copula constructions in other languages? I think everyone would consider those clauses. Also consider that nominative pronouns can appear in such constructions: ``I met a nice couple, \textbf{he a painter} and \textbf{she an engineer}.''---so these look like subjects.}\br{We're not doing away with the term ``verbless clause'', but they are only clauses in a semantic sense. CGEL is explicit that a) syntactic clauses are headed by VPs, and b) VlCs don't have a V or a gap where one would go. Note that CGEL would treat your example about the couple as a nonce-constituent, not a clause. Also nominative pronouns have functions beyond subject.}\nss{p. 1267: ``The underlined clauses have subject + predicate structure, but with no verb in the predicate.'' I don't know whether there should be a gap but it seems to me that if there is a predication, it must be a clause, verb or no verb. And I take it the distribution of complements in this construction is the same as or very similar to the distribution of complements of \w{be} (``(with) no crew present/*professionally/aboard/on the flight'') as opposed to \w{become} or \w{make} (*He became aboard, *They made him aboard). Note that VlCs can be negated with ``not'' (p. 811). Could we say VlCs are headed by VPs with fused head-complement? N.B. The above concerns verbless subordinate clauses that function similar to non-finite clauses. There are also verbless exclamatives (p. 921) and directives (p. 942) and comparatives (pp. 1111-12).}}

        \subsubsection{Names}
        Despite CGEL's claim that ``to a large extent the syntactic structure of proper names conforms to the rules for the structure of ordinary NPs,'' (p.~517), it also observes that ``there is no convincing evidence for treating one element as head'' in personal names (p.~519). Additionally, it notes that ``titles [of books, etc.]\ are much less constrained than other kinds of proper name and are excluded from [our] account'' (p.~517). For these reasons, we have chosen to treat proper names such as \pex{Pierre Vinken} as single lexical items. We similarly treat chemical compounds such as \pex{carbon dioxide} as single lexical items.
        
        In the dataset itself, multiple tokens in such a construction are analysed using the Flat relation, structured similarly to headless coordination.
        
        \subsubsection{Supplements}
        Supplements are adjuncts which are only loosely attached to a phrase, ``set off from the rest in speech by separate intonational phrasing and in writing by punctuation'' 
        (CGEL p.~25). A clear example is:
        
        \ex. \pex{The necklace, \uline{which her mother gave her \uline{~~~}}, was in the safe}
            \label{fig:supplement}
        
        There are, however, other cases where the distinction between supplements and modifiers is quite fuzzy. A fronted \pex{if}-PP, for instance, meets the prosodic and punctuational criteria of the definition. 
        But we currently have no practical and principled way to say whether the same PP at the end of a VP, for instance, is a modifier or a supplement. We have proceeded on intuition using discussions to try to resolve disagreements.
        
        Another issue is that the structural treatment of supplements in CGEL is sui-generis and inconsistent with typical trees. It consists of an anchor node to which a supplement such as the underlined relative clause in \cref{fig:supplement} is attached by means of a dashed arrow going from the supplement to the anchor. Under this analysis, the anchor and the supplements are not in a head-dependent relationship (p.~1350).
        For convenience and practicality, we instead attach supplements as dependents to the anchor head, even if that leads to three or more branches from the node, so that the NP in \cref{fig:supplement} would be [\pex{the necklace which her mother gave her}].
        
        \subsubsection{Miscellaneous}
        \begin{itemize}
            \item We treat comma splices as distinct clauses. 
            \item Clauses set off by punctuation such as commas, dashes, or (semi-)colons are analyzed as supplements.
            \item We take particles such as \pex{up} in \pex{pick up the kids} to be complements in a ternary branching VP along with the head and the object (CGEL p.~280).
            \item Ungrammatical aspects of the source sentence are kept and corrected using a special annotation.
        \end{itemize}
        

\section{Comparison with existing frameworks}\label{sec:comparison}

Having resolved some of the linguistic issues that arose while annotating a treebank in the CGEL formalism, we look towards comparing CGELBank with two established formalisms for computational syntax: Universal Dependencies \citep{nivre-20,de_marneffe-21} and Penn Treebank \citep{marcus-etal-1994-penn}.

\subsection{Corpus}

To enable quantitative comparisons, two selections of sentences were annotated as CGELBank trees:
\begin{itemize}
    \item \textbf{EWT}, 100~sentences from the English Web Treebank, which were previously annotated with Penn and UD trees \citep{ewt,silveira14gold}---we annotated CGELBank trees;
    \item \textbf{Ling}, 65~sentences observed to be linguistically interesting---we separately annotated CGELBank and UD trees.\footnote{UD trees were generated with Stanza v1.4.0 \citep{qi-etal-2020-stanza} and hand-corrected in the UD~Annotatrix tool \citep{tyers-etal-2017-ud}}\finalversion{\nss{explain the Twitter account}}
\end{itemize}
This resulted in a small set of parallel data. Summary statistics about our datasets are reported in \cref{tab:datasets}, and a breakdown by CGELBank labels (POS, phrasal category, grammatical function) in \cref{tab:catcounts}. CGELBank trees were originally created in \LaTeX{} using the \texttt{forest} package and later converted into a PENMAN-like textual notation \citep{mathiessen1991text}. Further details are in \cref{sec:selection}.

\subsection{POS tags}


\begin{table}
    \centering\small
    \setlength{\tabcolsep}{2.5pt}
    \adjustbox{max width=\columnwidth}{
    \begin{tabular}{rl rl rl}
    \toprule
    \multicolumn{2}{c}{\textbf{POS}} & \multicolumn{2}{c}{\textbf{Phrasal Category}} & \multicolumn{2}{c}{\textbf{Gram.~Function}} \\
    \cmidrule(r){1-2}\cmidrule(r){3-4}\cmidrule{5-6}

691 & N           & 1113 & Nom                  & 4444 & \func{Head} \\
361 & V           &  899 & NP                   &  627 & \func{Mod} \\
327 & P           &  807 & VP                   &  409 & \func{Comp} \\
303 & D           &  619 & Clause               &  403 & \func{Obj} \\
265 & N\textsubscript{pro}      &  343 & PP                   &  299 & \func{Det} \\
224 & V\textsubscript{aux}      &  302 & DP                   &  295 & \func{Subj} \\
173 & Adj         &  195 & AdjP                 &  209 & \func{Coordinate} \\
133 & Adv         &  134 & AdvP                 &  205 & \func{Marker} \\
104 & Sdr         &  103 & Coordination         &   88 & \func{PredComp} \\
101 & Coordinator &   86 & Clause\textsubscript{rel}          &   68 & \func{Supplement} \\
  4 & Int         &    6 & NP+PP                &   53 & \func{Flat} \\
   &  &   4 & IntP                 &   52 & \func{Det-Head} \\
    &             &    3 & NP+AdvP              &   43 & \func{Prenucleus} \\
96 & \textit{GAP}             &    2 & NP+Clause            &   10 & \func{Postnucleus} \\

        \bottomrule
    \end{tabular}}
    \caption{Counts in CGELBank POS tags, phrasal categories, and grammatical functions. Special phrasal categories for coordination and some functions are not listed due to low frequency.}
    \label{tab:catcounts}
\end{table}

The part-of-speech tagset used in CGELBank contains only 11 tags, fewer than the equivalent tagset for aligned tokens in both UD (18) and PTB (37).\footnote{CGELBank currently omits punctuation tokens, which account for some missing POS categories in UD and PTB. Of the substantive PTB tags we are missing only \texttt{WP\$} (\w{whose}).} 
The multilingual UD tagset is coarse due to the language-specific nature of fine-grained labels. The PTB tagset is fine-grained, reflecting morphological inflections of verbs, nouns, pronouns, adjectives, and adverbs, as well as WH status of pronouns, determiners, and adverbs. 
Even if the PTB tagset were coarsened, the correspondences across the three tagsets would not be exact.\finalversion{\nss{Suggests an experient: truncate \texttt{VB*} to \texttt{VB} so all verbs are grouped together, and likewise for nouns, adjectives, etc. Impact on entropy numbers?}}

\paragraph{Quantitive comparison.\footnote{Numbers reported henceforth were computed on a preliminary version of the CCGBank release (except for \cref{tab:catcounts}, which is current as of this writing). The comparison to PTB and UD identified inconsistencies in the CGEL trees that were subsequently revised. We report numbers prior to the revision because the revision was influenced by the comparison.}} Applying information theoretic measures to our data, we can estimate the overall uncertainty (i.e.~entropy) of the tagset, and whether the UD or PTB POS tag provides greater information about the CGELBank POS tag for a word (i.e.~conditional entropy given another distribution).
\begin{align}
    H(X) &= -\sum_{x \in \mathcal{X}}{p(x)\log{p(x)}}\\
    H(X \mid Y) &= -\sum_{x \in \mathcal{X}, y \in \mathcal{Y}}{p(x, y)\log{\frac{p(x, y)}{p(y)}}}
\end{align}
We induced an alignment of tokens between parallel trees (given that our hand-annotated CGELBank trees omitted punctuation, for example) and compared the aligned tokens' tags. 
As expected, when it comes to predicting the CGELBank tag, tags from other frameworks are informative, producing a marked reduction in entropy.
By this measure, the PTB tagset proves slightly more informative than UD. 

\begin{table}
    \centering
    \small
    \setlength{\tabcolsep}{3.5pt}
    \begin{tabular}{@{}lrrr@{}}
        \toprule
        & \textbf{EWT} & \textbf{Ling} & \textit{\textbf{Combined}} \\
        \midrule
        Trees & 99 & 65 & 164 \\
        Tokens & 2102 & 908 & 3010 \\
        \% \small{UD–CGELBk matched} & 86.6\% & 86.8\% & 86.6\%\\
        \midrule
        $H(\text{CGELBk POS})$ & 2.85 & 2.87 & 2.87\\
        $H(\text{CGELBk POS} \mid \text{UD POS})$ & 0.38 & 0.46 & 0.42 \\
        $H(\text{CGELBk POS} \mid \text{PTB POS})$ & \textbf{0.34} & \textbf{0.40} & \textbf{0.38}\\
        \bottomrule
    \end{tabular}
    \caption{Statistics for our data, including sizes and tag entropy values. Note: some UD tokens, such as punctuation, are omitted from CGELBank trees (unmatched).}
    \label{tab:datasets}
\end{table}

\paragraph{Differences.} The major contributors to tag uncertainty are the CGELBank categories of determinatives (D), prepositions (P), and subordinators (Sdr). These categories do not correspond neatly to the POS hierarchies defined by UD and PTB, leading to complex overlapping of the distribution of tags.


As summarized in \cref{tab:p_comparison}, CGEL adopts an expansive view of the preposition (P) category, incorporating function words that head traditional prepositional phrases as well as distributionally similar phrases (including intransitive particles, adverbials taking the form of a clause, deictic adverbial words, and so on). 
According to CGEL, characteristics unifying traditionally disparate categories as prepositions include that they are mostly closed-class and uninflecting; can mostly be modified by \w{right}; and can mostly appear adnominally, adverbially, and as complement of \w{be} (but not \w{become}).

As for determinatives (D), CGEL again takes a broad view of the situation, shown in \cref{tab:d_comparison}. The articles \pex{a} and \pex{the} are the prototypical determinatives. Beyond those, determinatives are lexemes that are mutually exclusive with the articles in NP structure (\pex{*the this book}) and/or can occur in the partitive construction (\pex{\uline{many} of them}) (p.~539). Unlike UD and PTB, determinatives (such as \w{this}) functioning as the sole constituent in the NP are not called pronouns; they are instead analyzed as fused determiner-heads of the NP (\cref{sec:treediffs}, \cref{itm:fused}).

\begin{table}
    \centering\small
    \setlength{\tabcolsep}{2pt}
    \adjustbox{max width=\columnwidth}{
    \begin{tabular}{@{}lccc@{}}
        & \textbf{CGELBank} & \textbf{PTB} & \textbf{UD} \\
        \midrule
        Relativizer \w{that} & Sdr & WDT & PRON \\
        Infinitive marker \w{to} & Sdr & TO & PART \\
        Complementizer \w{that}, \w{whether}, \w{if} & Sdr & IN & SCONJ \\
        \\
        Clause-marking \w{before}, \w{in}, \w{because}, \w{if}* \dots & P & IN & SCONJ \\
        Adjunct clause--marking \w{when}, \w{where} \dots & P & WRB & SCONJ\finalversion{**} \\
        NP-marking \w{of}, \w{before}, \w{in}, \w{to}, \w{out (of)} \dots & P & IN & ADP \\
        \\
        Idiomatic verb particle \w{out}, \w{up} \dots & P & RP & ADP \\
        Spatial particle \w{out}, \w{up} \dots & P & RB & ADV \\
        Adverbial \w{outside}, \w{here}, \w{then} \dots & P & RB & ADV \\
    \end{tabular}}
    \caption{Tagging of prepositions, subordinators, and related items in the three frameworks.\\ 
    *\,Non-complementizer (conditional) use of \w{if}\finalversion{; ** Current practice but may be changed to ADV in a future UD release}.}
    \label{tab:p_comparison}
\end{table}

Named entities are another source of tag divergence. CGELBank makes no distinction between proper nouns and common nouns, unlike the other formalisms. PTB goes as far as to label every content word in a proper name as NNP: e.g., \w{\uline{Supreme} Court} is labeled Adj by CGEL but NNP by PTB.

\begin{SCtable*}
    \centering\small
    \setlength{\tabcolsep}{2pt}
    \adjustbox{max width=1.5\columnwidth}{
    \begin{tabular}{@{}lccc@{}}
        & \textbf{CGELBank} & \textbf{PTB} & \textbf{UD} \\
        \midrule
        Article \w{a}, \w{an}, \w{the} & D & DT & DET \\
        Demonstrative \w{this}, \w{those} \dots & D & DT & DET, PRON \\
        \w{either}, \w{neither} & D & DT, CC, RB & DET, CCONJ, ADV \\
        \w{both} & D & DT, PDT, CC, RB & DET, CCONJ, ADV\\[4pt]
        Quantifier \w{all}  & D & DT, PDT & DET \\ 
        Quantifier \w{some}, \w{any}, \w{every}, \w{each} & D & DT  & DET \\ 
        Quantifier \w{few}, \w{little}, \w{many}, \w{much} & D & JJ & ADJ \\ 
        Quantifier \w{no} & D & DT, RB & DET, ADV \\[4pt] 
        \w{none} & D & NN & NOUN \\ 
        Nominal \w{anyone}, \w{nothing} \dots & D & NN & PRON \\
        Adverbial \w{anywhere}, \w{nowhere} & D & RB & ADV \\[4pt]
        \w{what} & Adj, D*, N\textsubscript{pro} & WP, WDT & PRON, DET \\
        \w{which} & D, N\textsubscript{pro} & WDT & PRON, DET \\[5pt]
        
        \w{one} & D, N, N\textsubscript{pro} & CD, NN, PRP & NUM, NOUN, PRON \\ 
        Other cardinal number, no suffix & D, N & CD & NUM \\
        \w{hundreds}, \w{thousands}, \dots & N & NNS & NOUN \\ 
    \end{tabular}}
    \caption{Selected CGELBank determinatives and related categories in other frameworks.\\ 
    *\, CGEL has \w{what} as a pronoun, determinative, or adjective (pp.~540, 909--910). The determinative use of \w{what} is not attested in CGELBank.
    \finalversion{\nss{`somewhere' is not entirely consistent in EWT: \url{https://github.com/UniversalDependencies/UD_English-EWT/issues/132}}}}
    \label{tab:d_comparison}
\end{SCtable*}


\subsection{Gaps}\label{sec:gap-empirical}
Following the discussion in \cref{sec:gap}, we sought to characterize CGELBank gaps empirically. We attempted to align them with PTB null\slash empty elements in the EWT set based on relative location between tokens in the sentence.
This reveals that CGELBank's gaps largely correspond to PTB elements \texttt{*T*} (e.g.~trace of WH-movement; 28 of 33 aligned) and \texttt{*RNR*} (Right Node Raising; 2 of 2 aligned).
There are 10 unaligned CGELBank gaps in EWT: these indicate noncanonical ordering of elements in a clause, whereas PTB signals this via a special constituent label---or not at all.
Conversely,
CGELBank does not use gaps for controlled subjects or other omitted subjects of nonfinite clauses (PTB \texttt{*PRO*}), nor for passivization.

\subsection{Tree structures}\label{sec:treediffs}


With regard to parse structure, CGELBank offers a systematic and coherent encoding of both phrase structure (as in PTB) and grammatical relations (as in UD). 
In this sense, CGELBank parses are more informative than either PTB or UD parses.

\paragraph{PTB.} Broadly speaking, CGELBank and Penn Treebank adopt comparable inventories of phrasal categories.\footnote{The main differences are that PTB uses finer-grained clause labels and assigns distinct categories to WH-phrases. CGEL defines a taxonomy of clause types, but CGELBank only distinguishes relative clauses (Clause\textsubscript{rel}) from others (Clause). More clause types may be added in the future.
} But the trees look very different for the following major reasons: 
\begin{enumerate}
    \item \textbf{Heads and Grammatical Functions:} PTB does not specify heads within phrases or systematically provide the grammatical function of a phrase with respect to its parent phrase.\footnote{In PTB, a limited set of functions are labeled on appropriate phrases, notably subjects and certain types of adjuncts.} Therefore, heuristic head rules \citep{collins} are necessary to convert to dependency structures.
    \item \textbf{Branching:} Constituents in the PTB are notoriously flat, with potentially multiple adjunct modifiers and complements all attaching at the same level. Often they are heuristically binarized for parsing.
    CGEL constituents are mostly binary already, with outer attachments for adjuncts, which makes the underlying grammar more parsimonious.
    \item \textbf{Empty Categories:} The PTB contains a variety of empty categories reflecting Government and Binding theory \citep{gb}.
    Our treebank contains a single notion of a gap (\cref{sec:gap}), limited primarily to unbounded dependency constructions and ellipsis; gaps do not represent control, for example. 
    \item \textbf{Fusion:} CGEL's fused-head construction occurs when the \func{Head} function is fused with a dependent function in NP structure (p.~332; more in ch.~5§9). The \func{Prenucleus:}NP in \cref{fig:5brevised} illustrates one of three such fusions (see \cref{fig:FunctionSet}). The fused-head construction results in a directed acyclic graph \citep{pullum2008expressive}.
    In these cases, the dependent simultaneously performs its usual function (e.g., \func{Det}:DP) and the function of \func{Head} in the NP, leading it to have two parent nodes in the graph. We maintained the tree structure in our dataset such that the fused-head relations were still recoverable; only one edge was kept but specially labelled as \func{$x$-Head} or \func{Head-$x$}.
    
    PTB has no counterpart to CGEL's fusion of functions. All PTB parses are properly trees. 
    \label{itm:fused}
\end{enumerate}

\paragraph{UD.} After aligning tokens between UD and CGELBank representations, we find that only 46.6\% of them have the same head\footnote{To extract heads of tokens from CGELBank trees, we treat both head and fused-head constituents as heads of their sister constituents. In flat structures (e.g.~coordination) we take the first element as the head to agree with UD's representation.} in both formalisms. This is largely due to CGEL's treatment of many function words as heads, since e.g.~a preposition projects a prepositional phrase, not a nominal phrase.
Consequently, the proportion of UD dependencies with CGELBank equivalents is low for auxiliaries (\deprel{aux}, 4.1\% head agreement), copulas (\deprel{cop}, 0.0\%), and standard prepositions (\deprel{case}, 0.0\%). 

UD's design decision to favor content heads was driven by valid crosslinguistic considerations but leads to awkward treatment of English prepositions and copular constructions \cite{gerdes2018sud}.\footnote{For example, consider \w{The problem is that the project will soon be behind schedule.} In CGEL, the head of the clause is \w{is}, with a \w{that}-clause as complement, etc. In UD, the main predicate is \w{schedule} (since copulas, other auxiliaries, complementizers, and prepositions are considered modifiers).}
A second consequence of content heads (combined with the lack of constituent structure) is that some structural ambiguities are not resolved by the parse, as in the case of coordination: clause-level, verb phrase--level, and predicate-level coordination produce identical dependency structures.\footnote{E.g., [\pex{Quickly dress}] \pex{and} [\pex{cook the chicken}] vs.~\pex{Quickly} [[\pex{dress}] \pex{and} [\pex{cook the chicken}]] vs.~\pex{Quickly} [\pex{dress and cook}] \pex{the chicken}. For all three interpretations, the UD tree will have \w{dress} as the clause root with \w{cook} (\deprel{conj}) and \w{quickly} (\deprel{advmod}) as its dependents, and no indication \w{dress} and \w{cook} can simultaneously be transitive with a shared direct object.} 

Several dependency relations in UD accommodate ``extrasyntactic'' relationships that appear in corpus sentences, such as repair and headless expressions (e.g., personal names). But when it comes to compositional syntactic relations, CGELBank's inventory of phrasal categories and functions is finer-grained and generally more informative.

One example is complementation: English UD only distinguishes complements from modifiers if they are \textit{non-prepositional} dependents within a \textit{clause}---subjects, objects, and complement clauses in UD are not prepositionally marked and can only be dependents of a clausal predicate. 
Thus, all clausal complements of nouns (\pex{the claim \uline{that there is life on Mars}}), as well as prepositional (oblique) clausal complements of any word (\pex{interested \uline{in living on Mars}}), are labeled the same as modifier clauses in UD \citep{przepiorkowski-19}.

Sometimes UD basic (surface-level) structures provide more information than CGELBank: 
\begin{enumerate}
    \item The \deprel{xcomp} relation (borrowed from LFG; \citealp{dalrymple2001lexical}) is used within a clause to signal predicate-licensed control or raising. Control and raising are not indicated in CGELBank.
    \item The UD \deprel{expl} relation marks expletive uses of \w{it} and \w{there} (\pex{\uline{It} is clear that \uline{there} is more work to do}), signaling that these words are nonreferential and present only for syntactic reasons.
    In CGELBank, these are treated as pronouns filling ordinary grammatical functions (usually \func{Subj})---though some constructions that trigger expletive pronouns, like extraposition (\cref{sec:extraposition}), are specially marked elsewhere in the tree.
    \item The subject relations in English UD distinguish passive from non-passive subjects. CGEL analyzes the passive construction as combining ordinary morphosyntactic components at the surface level, so no special label is present in the tree.
    \item English UD has special relations for nominal expressions serving as temporal modifiers. Arguably this is a place where semantics has crept into the syntax.
    \item As UD requires every word to be assigned a dependency relation, treebanks are required to make decisions about the internal structure of expressions like dates, though these are not well standardized at present \citep{zeman-21,schneider-21}.
\end{enumerate}

In addition, several UD languages, including English, offer \emph{enhanced} dependency graphs that augment the surface-syntactic tree with deeper relations \citep{englishud-enhanced}.
CGELBank reflects mostly ``surface structure'', though its coindexation of gaps can be used to recover some of the UD enhanced dependencies (e.g., for relative constructions).

\section{Conclusion}

Using the analysis developed in CGEL \citep{cgel}, we introduced a new expressive and linguistically-informed syntactic formalism to corpus annotation of English. We released a small treebank with parallel annotation in CGELBank, UD, and PTB, and compared features of the three. The unique features of the CGEL formalism, combined with its minimal complexity, lend themselves to computational work on English syntax.

In the future, we plan to expand our treebank with more manually-annotated trees, using that to develop means for automatic conversion from UD and/or PTB to CGELBank. A full annotation manual is under development. We also are interested in incorporating more of CGEL's analysis, particularly in morphology and fine-grained part-of-speech tags.

\section*{Acknowledgments}
We thank John Payne, Geoffrey Pullum, and Pairoj Kunanupatham for useful discussion on many linguistic issues that arose in annotation as well as the CGEL formalism in general. We also thank various Twitter users (including Rui P. Chaves and Russell Lee-Goldman) who commented on syntactic analyses posted by Brett Reynolds, as well as anonymous reviewers.

\bibliography{anthology,custom}
\bibliographystyle{acl_natbib}

\newpage
\appendix

\section{Functions}\label{sec:functions}
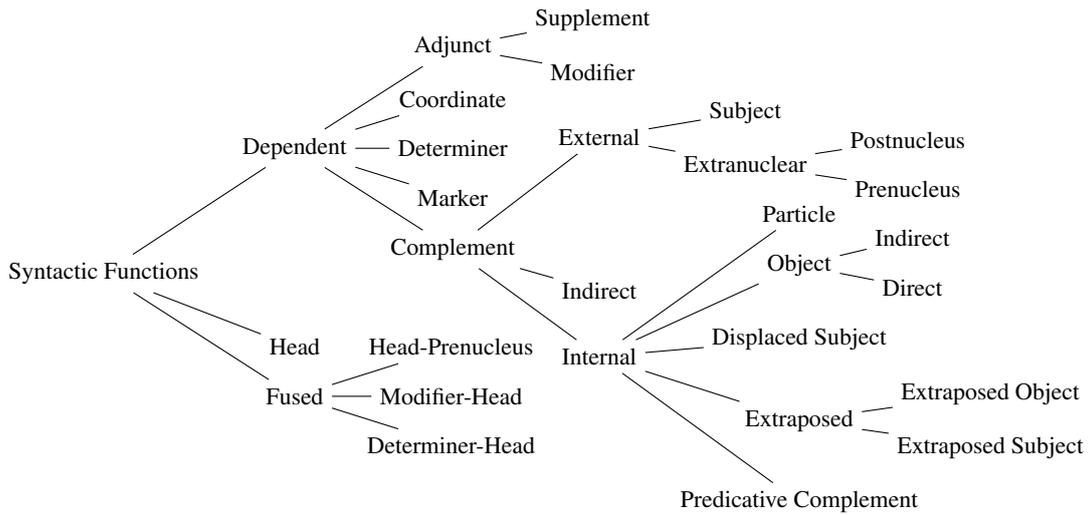
\begin{figure*}
        \small
        \centering
        \begin{forest}
        [Syntactic Functions, grow=east
			[Fused, grow=east
				[Determiner-Head]
				[Modifier-Head]
				[Head-Prenucleus]
			]
			[Head, grow=east]
			[Dependent, grow=east
				[Complement, grow=east
					[Internal, grow=east
						[Predicative Complement]
						[Extraposed, grow=east
							[Extraposed Subject]
							[Extraposed Object]
						]
						[Displaced Subject]
						[Object, grow=east
							[Direct]
							[Indirect]
						]
						[Particle]
					]
					[Indirect]
					[External, grow=east
						[Extranuclear, grow=east
							[Prenucleus]
							[Postnucleus]
						]
						[Subject]
					]
				]
				[Marker]
				[Determiner]
				[Coordinate]
				[Adjunct, grow=east
					[Modifier, grow=east]
						[Supplement]
				]
			]
		]
        \end{forest}
        \caption{The hierarchical relation of all functions in the treebank.\finalversion{\nss{make sure this is up to date}}}
        \label{fig:FunctionSet}
    \end{figure*}

    The hierarchy of the full set of functions used in the treebank is shown in Figure \ref{fig:FunctionSet}.

\section{Corpus design}\label{sec:selection}

\paragraph{EWT.} We randomly sampled 200 trees from the portion of sentences in EWT that have token count between 15 and 30 inclusive. These sentences were long enough to naturally contain complex linguistic phenomena that could test our annotation guidelines, and preliminary attempts at rule-based conversion from UD to CGELBank found difficulties in sentences in this length range. 99 of the trees were manually annotated thus far by one of the authors. Another author performed extensive consistency checks and revisions, including splitting one of the sentences into two sentences, resulting in 100~annotated trees.

\paragraph{Ling.} An author encountered linguistically-interesting and difficult-to-analyze sentences in a variety of contexts (news articles/headlines, social media posts, etc.)\ over the course of a year and manually annotated them in the CGEL formalism. These were annotated before some of the guidelines in this paper were agreed upon, leading to many necessary revisions. 

\section{More tables}
\Cref{tab:ambig_classes} lists lemmata with ambiguous POS tags in CGELBank. \Cref{tab:fxnwords} lists all function words found in CGELBank.

\begin{table}
    \centering\small
    \setlength{\tabcolsep}{2pt}
    \begin{tabular}{@{}>{\raggedright}p{7em}p{7.5em}p{9em}@{}}
    \toprule
\{D, Sdr\}: \w{that} & \{P, V, Adj\}: \w{like} & \{N, V\}: \w{call} \w{try} \\
 
\{D, N\textsubscript{pro}\}: \w{one} & \{P, N\textsubscript{pro}\}: \w{there} & \{V, V\textsubscript{aux}\}: \w{do} \w{have} \\
 \{D, Adv\}: \w{more} & \{P, Adv\}: \w{as} & \{Adj, Adv\}: \w{very} \w{well} \\
 & \{P, Sdr\}: \w{to} \w{for} \w{if} & \{Adj, N\textsubscript{pro}\}: \w{what} \\
& \multicolumn{2}{p{13em}}{\{P, Coordinator, Int, Adv\}: \w{so}}\\
\bottomrule
    \end{tabular}
    \caption{Ambiguous lemmata occurring at least 5 times}
    \label{tab:ambig_classes}
\end{table}

\begin{table*}
    \centering\small
    \begin{tabular}{@{}>{\raggedright}p{25em}@{}p{0pt}@{}}
        \toprule
        \textbf{D} [excluding numerals]: a, a few, a little, all, another, any, anybody, anyone, anything, anywhere, both, each, enough, every, everybody, everyone, everything, least, many, many a, million, more, no, no one, none, one, several, some, someone, something, sometimes, somewhere, that, the, this, three, two, which & \\
        \midrule
        \textbf{N\textsubscript{pro}:} he, I, it, its, mine, my, one, she, there, they, we, what, which, who, yesterday, you, your & \\
        \midrule
        \textbf{P:} @, a.m., about, above, after, against, along, around, as, at, away, back, because, before, behind, between, by, considering, coupled, down, due, during, for, forward, from, here, if, in, in order, including, into, irrespective, like, now, of, off, on, onboard, out, outside, over, past, per, regarding, since, so, so long as, than, then, there, through, to, up, upon, upstairs, when, while, with, within & \\
        \midrule
        \textbf{V\textsubscript{aux}:} be, can, could, do, have, may, must, should, will, would & \\
        \midrule
        \textbf{Sdr:} for, if, that, to, whether & \\
        \midrule
        \textbf{Coordinator:} \&, -, /, and, but, etc, or, plus, so & \\
        \bottomrule
    \end{tabular}
    \caption{Full list of function word lemmata in our data by POS}
    \label{tab:fxnwords}
\end{table*}

\begin{figure*}
    \centering
    \includegraphics[width=6.5cm]{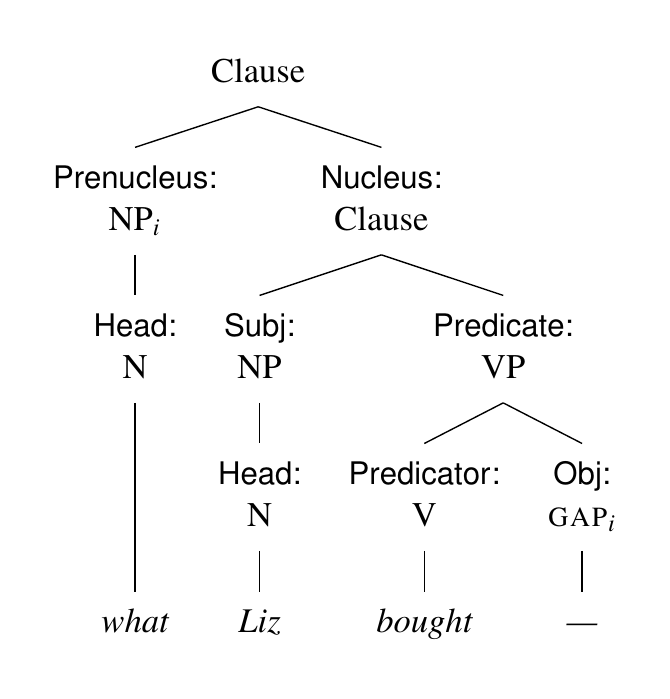}
    \caption{The interrogative clause \pex{what Liz bought} as originally parsed in CGEL (p.~48, figure 5b).}
    \label{fig:5b}
\end{figure*}

\end{document}